\documentclass[12pt]{article}
 \usepackage{authblk}
 \usepackage[utf8]{inputenc} 
\usepackage[T1]{fontenc}    
\usepackage{hyperref}       
\usepackage{url}            
\usepackage{booktabs}       
\usepackage{amsfonts}       
\usepackage{nicefrac}       
\usepackage{microtype}      
\usepackage{lipsum}
\usepackage{graphicx}
\usepackage{subfig} 
\usepackage{caption}
\usepackage{multirow}
\usepackage{multicol}
\usepackage[margin=1in]{geometry}
\usepackage{setspace}
\usepackage{footnote}
\usepackage{diagbox} 
\usepackage{float}
\usepackage{booktabs}
\usepackage{threeparttable}
\usepackage{booktabs} 
\usepackage[linesnumbered,ruled]{algorithm2e}
\usepackage{amsmath,amsthm,amssymb}
\usepackage{color}

\usepackage[margin=1in]{geometry} 
\usepackage{amsmath,amsthm,amssymb}
\usepackage[round]{natbib}

 \usepackage{amsmath}

 \usepackage{amsmath}
\usepackage[utf8]{inputenc}
\usepackage{hyperref}

\usepackage{multirow} 
\usepackage{multicol} 
\usepackage{arydshln}
 \usepackage{hyperref}

\author[1]{Lin Xiao}
\author[1]{Luo Xiao}
\affil[1,2]{North Carolina State University}
\date{}

\title{Linked Component Analysis for Multiview Data}

\begin{document}
 

\maketitle

\section{Introduction}
Recent technological advances have led to increased availability of multiple sources of high-content data. In particular, multiview data refers to  different types of variables collected from the same set of individuals. One typical example is the Roadmap Epigenomics Project \citep{kundaje2015integrative} which integrates information about histone marks, DNA methylation, DNA accessibility and RNA expression to infer high-resolution maps of regulatory elements annotated jointly across a total of 127 reference epigenomes spanning diverse cell and tissue types. Another example is the data used in NCI-DREAM drug sensitivity prediction challenge (\cite{costello2014community}) which contains 
gene expression (GE), RNA, DNA methylation (MET), copy number variation (CNV),
protein abundance (RPPA) and exome sequence (EX) measurements for 53 human breast
cancer cell lines. The prevalence of multiview data has motivated research on uncovering associations between different data views. For instance, many studies in neuroscience have been focusing on identifying joint relationship between nonimaging features including genetic, demographics, behavioral, clinical information and imaging derived features
 represented by brain structural measurements, functional connectivity measurements and electrophysiological recordings \citep{zhuang2020technical}. This type of association analysis can hopefully provide insights on the disease-related links between neurobiological activities and phenotypic features.
 
In application, canonical correlation analysis (CCA) is one of the most commonly used methods for extracting common variation shared among two data views. CCA aims at seeking for a pair of canonical coefficients such that the correlation between two transformed data views is maximized. From the perspective of a latent factor model, CCA assumes that all data views are generated from a common latent subspace of which the rank is lower than that of any data view and the aim is to identify the common latent subspace. A number of CCA variants have been proposed to adapt to more complex scenarios. In order to capture the potentially nonlinear correlation between different data views, kernel CCA (\cite{akaho2006kernel}; \cite{melzer2001nonlinear}) and deep CCA \citep{andrew2013deep} are proposed, where kernel CCA projects each data view onto a new feature space through a nonlinear mapping embedded in a fixed kernel function while deep CCA learns this nonlinear representation via deep neural networks in a more flexible manner. In a high dimensional setting where the number of features might exceed the number of observations, sparse CCA (\cite{witten2009penalized}; \cite{wang2019identify}) induces sparsity on the canonical coefficients for the purpose of providing more reliable and interpretable estimation. Classic CCA-based methods are usually developed for two data views, though, extension to the situation where multiple views are available is quite straightforward. There are a variety of formulations for multiset CCA and they differ in the proposed optimization objective functions. Among these formulations, some propose to maximize the pairwise correlations or pairwise covariances (SUMCOR CCA, SSQCOR CCA, SABSCOR CCA) and the others try to directly identify a common subspace such that the Euclidean distance between the common subspace and the projected data view is maximized (MAXVAR CCA, LS-CCA). We refer to \cite{kettenring1971canonical} for a more comprehensive summary of possible multiset CCA formulations. Another related work \citep{min2020sparse} introduces the multiple co-inertia analysis (mCIA) method, which aims to find a set of co-inertia loadings and a synthetic center such that the weighted sum of squared covariances between each transformed data view and the synthetic center is maximized. 

While the aforementioned methods primarily  focus on common relationships across different data views, many existing approaches take a richer set of association structures into account. For instance, JIVE \citep{lock2013joint}, COBE \citep{zhou2015group}, and AJIVE \citep{feng2018angle} explore joint and individual structures simultaneously. Going further, \cite{jia2010factorized}, \cite{van2011flexible} and SLIDE \citep{gaynanova2019structural} take partially shared structures into consideration. The majority of these methods take advantage of matrix decomposition models. Specifically, each data view is decomposed into the summation of a signal matrix and an error matrix, and the signal part is further decomposed into several low-rank terms with each term representing a specific association structure. In particular, the joint structure here is typically defined as the intersection of the column spaces of the signal matrices associated with each data view. 

Among the aformentioned approaches, CCA-based and CIA-based methods typically extract the shared subspace in a successive way. A natural question that would arise for these sequential methods is how to determine the number of shared components, or equivalently, how to estimate the rank of the common latent subspace shared by all data views. In a situation where two data views are available, rank selection can be performed based on the cumulative correlation plot or the scree plot for the sample cross covariance matrix \citep{kim2021seedcca}. Specifically, this sequential procedure can be terminated at the point where the increment in the cumulative correlation or the decline in the singular values becomes insignificant, or where the pre-specified percentage of cumulative canonical correlation or cumulative variation has been met. Despite the handiness of these empirical rules, one obvious drawback is that they require manual inspection and hence the resulting decision is more or less subjective. As for the matrix-decomposition based approaches, the majority of works focus on estimation of low rank components associated with each data view and a direct approach for rank identification is missing except JIVE and SLIDE. JIVE provides two rank selection schemes when the underlying rank is unknown: the rank of the joint signal is determined either through a permutation test procedure or via a Bayesian Information Criterion (BIC) selection algorithm. In addition, SLIDE provides estimation of the loading matrices and score matrices associated with each type of structure and the rank for each structure including the joint structure can be directly identified by counting the number of the columns of the corresponding loading matrix.

 In this work, we propose the joint linked component analysis (joint\_LCA) for multiview data. Unlike classic methods which extract the shared components in a sequential manner, the objective of joint\_LCA is to identify the view-specific loading matrices and the rank of the common latent subspace simultaneously. We formulate a matrix decomposition model where a joint structure and an individual structure are present in each data view, which enables us to arrive at a clean svd representation for the cross covariance between any pair of data views. An objective function with a novel penalty term is then proposed to achieve simultaneous estimation and rank selection. In addition, a refitting procedure is employed as a remedy to reduce the shrinkage bias caused by the penalization.
 
The paper is organized as follows. In section \ref{sec:methods}, we introduce the matrix decomposition model and formulate the objective function for joint\_LCA. In section \ref{sec:estimation}, we elaborate on the algorithm used for solving the optimization problem, along with the initialization schema and the refitting procedure. In section \ref{sec:simulation}, we compare the performance of joint\_LCA with other competitors in terms of estimation error and rank selection accuracy through extensive simulations. In section \ref{sec:real_data}, we apply joint\_LCA for integrative analysis of
real multiview data. We conclude with discussion in section \ref{sec:discussion}.


\section{Proposed Method} \label{sec:methods}
Suppose that we have $I$ data views from $n$ subjects, i.e, $X_i\in \mathbb{R}^{n\times p_i}$ for $i=1,\cdots,I$ and the sample cross covariance matrix between data view $X_i$ and data view $X_j$ is denoted as $\hat{\boldsymbol{\Sigma}}_{ij}$. The data generative model is:
\begin{equation*}
    X_i = U(V_iD_i)^{\intercal} + U_{i0}(V_{i0}D_{i0})^{\intercal} + \boldsymbol{W}_i,\quad i=1,\cdots, I
\end{equation*}
where the $U(V_iD_i)^{\intercal}$ is the submatrix of the joint structure that is associated with $X_i$, and $U_{i0}(V_{i0}D_{i0})^{\intercal}$ represents the individual structure unique to $X_i$. The score matrix $U\in \mathbb{R}^{n\times p_0}$ is shared across all data views, and score matrices $U_i$ is specific to $X_i$. $D_i\in \mathbb{R}^{r_0\times r_0}$ is a diagonal matrix with singular values corresponding to the joint structure associated with $X_i$. $r_0$ denotes the rank of the common structure among all data views, namely, the number of common components shared across $X_is$. $W_i$ is the error matrix. We assume that the columns of score matrices $U$ and $U_{i0}s$ are independent random vectors with mean zero and variance 1. Hence, if we calculate the cross covariance matrix between any two data matrices $X_i$ and $X_j$:
\begin{equation}
\label{eq:cc}
\begin{split}
    \text{cov}(X_i,X_j) &= \text{E}\left[X_i^{\intercal}X_j\right]\\
    &=V_iD_i\text{E}\left[U^{\intercal}U\right]D_j^{\intercal}V_j^{\intercal}+V_iD_i\text{E}\left[U^{\intercal}U_{j0}\right]D_{j0}^{\intercal}V_{j0}^{\intercal}\\
    &+V_{i0}D_{i0}\text{E}\left[U_{i0}^{\intercal}U\right]D_{j}^{\intercal}V_{j}^{\intercal}+V_{i0}D_{i0}\text{E}\left[U_{i0}^{\intercal}U_{j0}\right]D_{j0}^{\intercal}V_{j0}^{\intercal}\\
    &=V_iD_iD_j^{\intercal}V_j^{\intercal}
\end{split}    
\end{equation}
The last equality is due to the fact that
$\text{E}\left[U^{\intercal}U\right]=\boldsymbol{I}_n$, $\text{E}\left[U_{i0}^{\intercal}U\right]=\boldsymbol{0}$, $\text{E}\left[U^{\intercal}U_{j0}\right]=0$, and $\text{E}\left[U_{i0}^{\intercal}U_{j0}\right]=0$.

Unlike classical canonical correlation methods with sequential formation,
the goal of our method is to jointly estimate the number of common factors $r_0$ and the canonical loading matrices $V_i,i=1,\cdots,I$. Based on our data generative model and the cross covariance \eqref{eq:cc}, we propose the following optimization problem:
\begin{equation}
\begin{split}
    \min_{U_i,D_i}  &\sum_{1\leq<i<j\leq I} w_{ij} \|\hat{S}_{ij}-V_iD_iD_j^{\intercal}V_j^{\intercal}\|_F^2 +\lambda\sum_{k=1}^{p_0} \sqrt{\sum_{1\leq i<j\leq I }w_{ij}\sigma_{ijk}^2}\\
    &s.t. \quad V_i^{\intercal}V_i = I \quad \forall 1\leq i \leq I
\end{split}
\end{equation}\label{eq:model1}

The data fidelity term in our objective function is a weighted sum of the squared loss $\|\hat{S}_{ij}-V_iD_iD_jV_j^{\intercal}\|_F^2$. In order to perform rank selection simultaneously, we add a penalty term as follows:
\begin{equation*}
    \sum_{k=1}^{p_0} \sqrt{\sum_{1\leq i<j\leq I }\sigma_{ijk}^2}
\end{equation*}
where $\sigma_{ijk}=d_{ik}d_{jk}$ and $d_{ik}$ is the $k$th diagonal entry of $D_i$. This penalty enforces all $D_i, i=1,\cdots,I$ to have the same rank, which is exactly the number of common factors. $w_{ij}$ is a positive weight associated with the canonical correlation between data matrix $X_i$ and $X_j$. In specific, $w_{ij}$ can be constructed as $1/\| \hat{S}_{ij}\|_F^2$, the inverse of the squared Frobenius norm of the sample cross covariance matrix $\hat{S}_{ij}$. And we choose $p_0 = \min_{1\leq i \leq I}p_i$, which is the maximal possible number of common components. (The rank of solutions is penalized by adding a group penalty on each diagonal position across
all views)

\section{Estimation} \label{sec:estimation}
\subsection{Algorithm}
In this section, we describe how to solve optimization problem (\ref{eq:model1}). The minimization of the objective function with respect to parameters $\left\{(V_i,D_j)\right\}_{i=1}$ is approached by iteratively updating each of the parameters ($V_i$ or $D_i$) in turn, keeping others fixed. 

First, for finding a solution of $V_1$, we fix $D_i, i=1,\cdots,I$ and $V_i, i=2,\cdots,I$. Then problem (\ref{eq:model1}) becomes 
\begin{equation}
\label{eq:U}
    \begin{split}
    \min_{V_1} \sum_{1<j\leq I} w_{1j} \|\hat{S}_{1j}-V_1D_1D_j^{\intercal}V_j^{\intercal}\|_F^2 \\
    \text{s.t.} \quad V_1^{\intercal}V_1 = I
    \end{split}
\end{equation}
Noting that the objective function in \eqref{eq:U} can be rewritten as 
\begin{equation*}
    \|\hat{S}_{1,-1} - V_1M_{1,-1}\|_F^2
\end{equation*}
where $\hat{S}_{1,-1}$ is the concatenation of all pairwise cross covariance matrices between data view $X_1$ and any other data view, that is, $\hat{S}_{1,-1}=\left[\sqrt{w_{12}}\hat{S}_{12},\cdots,\sqrt{w_{1I}}\hat{S}_{1I}\right]$. And $M_{1,-1} = \left[\sqrt{w_{12}}D_1D_2V_2^{\intercal},\cdots,\sqrt{w_{1I}}D_1D_IV_I^{\intercal}\right]$ is the concatenated matrix of $D_1D_jV_j^{\intercal}$ for all $j\neq 1$.
Hence, the problem (\ref{eq:U}) can be transformed to the well-known orthogonal procrustes problem below:
\begin{equation*}
    \begin{split}
    \min_{V_1} \|\hat{S}_{1,-1} - V_1M_{1,-1}\|_F^2\\
      \text{s.t.} \quad V_1^{\intercal}V_1 = I
    \end{split}
\end{equation*}
which has a closed form solution $V_1 = RQ^{\intercal}$, where $R$ and $Q$ are the left and right singular vectors of $\hat{S}_{1,-1}M_{1,-1}^{\intercal}$, and can be obtained by performing SVD on $\hat{S}_{1,-1}M_{1,-1}^{\intercal}$.
Rest of the loading matrices $V_i, i=2,\cdots,I$ can be estimated by applying the same procedure as $V_1$.

Next, we consider finding a solution to $D_i, i=1,\cdots,I$.  We first simplify the data fidelity term in equation (\ref{eq:model1}). Taking into account the fact that $V_i, i=1,\cdots,I$ are orthonormal matrices, we have
\begin{equation*}
    \begin{split}
&\sum_{1\leq i<j \leq I}w_{ij}\|\hat{S}_{ij}-V_iD_iD_j^{\intercal}V_j^{\intercal} \|_F^2 \\
&=\sum_{1\leq i<j \leq I}w_{ij}\|V_i^{\intercal}\hat{S}_{ij}V_j-D_iD_j^{\intercal} \|_F^2 \\
&= \sum_{1\leq i<j \leq I} w_{ij}\|\tilde{S}_{ij}-D_iD_j^{\intercal}\|_F^2
\end{split}
\end{equation*}
where $V_i^{\intercal}\hat{S}_{ij}V_j$ is denoted as $\tilde{S}_{ij}$. Since $D_i, i=1,\cdots,I$ is diagonal, the above term can be further simplified as follows:

\begin{equation*}
    \begin{split}
        &\sum_{1\leq i <j \leq I}w_{ij}\|\tilde{S}_{ij}-D_iD_j^{\intercal} \|_F^2\\
        &= \sum_{1\leq i<j \leq I}w_{ij}\sum_{k=1}^{p_0}\left((\tilde{S}_{ij})_{kk}-d_{ik}d_{jk}\right)^2 + C\\
        &=\sum_{1\leq i<j \leq I}w_{ij}\sum_{k=1}^{p_0}\left((\tilde{S}_{ij})_{kk}-\sigma_{ijk}\right)^2+C
    \end{split}
\end{equation*}
where $(\tilde{S}_{ij})_{kk}$ denotes the $k$th diagonal entry of $\tilde{S}_{ij}$ and $C$ is a constant value that does not depend on $D_i,i=1,\cdots,I$, which can be removed from the objective function when estimating $D_i, i=1,\cdots,I$.

Now we combine the simplified data fidelity term and the penalty term and obtain the following objective function:
\begin{equation*}
    \min_{D_i,i=1,\cdots,I}\sum_{k=1}^{p_0}\sum_{1\leq i<j \leq I}w_{ij}\left((\tilde{S}_{ij})_{kk}-\sigma_{ijk}\right)^2+\lambda\sum_{k=1}^{p_0}\sqrt{\sum_{1\leq i<j \leq I} w_{ij}\sigma_{ijk}^2}
\end{equation*}

Apparently, the above objective function is decomposable with respect to index $k=1,\cdots,p_0$. Hence, we can obtain solutions for $\boldsymbol{\sigma}_k = (\sigma_{ijk})_{1\leq i <j \leq I}$ by solving the optimization problem below for each fixed $k$:
\begin{equation}
\label{eq:sigma}
    \min_{\sigma_{ijk}}\sum_{1\leq i<j \leq I}\left(\sqrt{w_{ij}}(\tilde{S}_{ij})_{kk}-\sqrt{w_{ij}}\sigma_{ijk}\right)^2+\lambda\sqrt{\sum_{1\leq i<j \leq I} w_{ij}\sigma_{ijk}^2} 
\end{equation}
(\ref{eq:sigma}) can be viewed as a problem with the format 'squared loss'+'group-lasso' penalty with respect to $\tilde{\boldsymbol{\sigma}}_k=(\sqrt{w_{ij}}\sigma_{ijk})_{1\leq i<j \leq I}$, and hence the solution can be explicitly expressed as 
\begin{equation*}
   \tilde{\boldsymbol{\sigma}}_k =\left( 1-\frac{\lambda}{\boldsymbol{Y}_{k}} \right)_{+}\boldsymbol{Y}_{k}
\end{equation*}
where $\boldsymbol{Y}_k = \left(\sqrt{w_{ij}}(\tilde{S}_{ij})_{kk}\right)_{1\leq i< j\leq I}$. Then $\boldsymbol{\sigma}_k$ can be derived immediately from $\tilde{\boldsymbol{\sigma}}_k$. 

Based on solutions for $\boldsymbol{\sigma}_k$, denoted as $\hat{\boldsymbol{\sigma}}_k$, we propose to estimate the diagonal entries $d_{ik},k=1,\cdots,p_0$ of $D_i, i=1,\cdots,I$ by solving the following least-square problem:
\begin{equation}
\label{eq:dk}
    \begin{split}
       \min_{d_{ik}} \sum_{k=1}^{p_0}\sum_{1\leq i < j \leq I}\left(d_{ik}d_{jk}-\hat{\sigma}_{ijk}\right)^2\\
        s.t. \quad d_{ik}\geq 0, \forall i=1,\cdots,I,  k=1,\cdots,p_0
    \end{split}
\end{equation}
Similarly, this problem is equivalent to solve 
\begin{equation}
\label{eq:d}
     \begin{split}
       \min_{d_{ik},i=1,\cdots,I} \sum_{1\leq i < j \leq I}\left(d_{ik}d_{jk}-\hat{\sigma}_{ijk}\right)^2\\
        s.t. \quad d_{ik}\geq 0, \forall i=1,\cdots,I
    \end{split}
\end{equation}
separately/simultaneously for each $k$, which can also be computed by an iterative process. 
Suppose that $d_{jk},j=2,\cdots, I$ are fixed, and we want to estimate $d_{1k}$ by solving
\begin{equation}
\label{eq:d1}
    \begin{split}
       \min_{d_{1k}} \sum_{ 1< j \leq I}\left(d_{1k}d_{jk}-\hat{\sigma}_{1jk}\right)^2\\
        s.t. \quad d_{1k} \geq 0
    \end{split}
\end{equation}
It is straightforward to derive the solution for $d_{1k}$:
\begin{equation}
\label{eq:d11}
 d_{1k}=\begin{cases}
 \frac{\sum_{1<j\leq I}d_{jk}\hat{\sigma}_{1jk}}{\sum_{1<j\leq I}d_{jk}^2} , & \sum_{1<j\leq I}d_{jk}\hat{\sigma}_{1jk}>0 \quad \text{and} \quad \sum_{1<j\leq I}d_{jk}^2\neq 0\\
 0 &\sum_{1<j\leq I}d_{jk}\hat{\sigma}_{1jk}<0 \quad \text{or} \quad\sum_{1<j\leq I}d_{jk}^2= 0
 \end{cases}
\end{equation}
Other variables $d_{jk},j=2,\cdots,I$ can be updated iteratively in a similar manner.

\subsection{Initialization}

As problem (\ref{eq:model1}) is not convex, convergence to the global optimum is not guaranteed, and the solution would depend on starting values $\left\{V_i^{(0)},D_i^{(0)} \right\}_{i=1}^{I}$. As initial values in optimization, we use an approximation of the solution under the assumptions
that all components are globally joint. Specifically, we initialize $V_i^{(0)}$ with the first $p_0$ left singular vectors of $S_{i,-i}$, which is the concatenation of all pairwise cross covariance matrices between data view $X_i$ and other matrices. In terms of initializing $D_i,i=1,\cdots,I$, We propose to use the solutions to the optimization problem below as the starting values for the diagonal entries $d_{ik}, i=1,\cdots,I, k=1,\cdots,p_0$:
\begin{equation*}
    \begin{split}
       \min_{d_{ik},i=1,\cdots,I,k=1,\cdots,p_0} \sum_{k=1}^{p_0}\sum_{1\leq i < j\leq I } \left[d_{ik}d_{jk}-\max\left(\left[(V_i^{(0)})^{\intercal}\hat{S}_{ij}V_j^{(0)}\right]_{kk},0\right)\right]^2
    \end{split}
\end{equation*}
Assuming that for a fixed $k$, all $d_{ik},i=1,\cdots,I$ are equal, then the solution to the above problem would be
\begin{equation*}
    d_{1k}=\cdots=d_{Ik}=\sqrt{\frac{\sum_{1\leq i<j\leq I}\max\left(\left[(V_i^{(0)})^{\intercal}\hat{S}_{ij}V_j^{(0)}\right]_{kk},0\right)}{I(I-1)/2}}, \forall k=1,\cdots,p_0
\end{equation*}

\subsection{Model Refitting Given a Pre-specified Rank}
By fitting \eqref{eq:model1}, the number of nonzero diagonal entries of $D_i,i=1,\cdots,I$, denoted as $r$, is our estimate of the rank of common components. However, the penalty term would cause shrinkage on the nonzero diagonal entries at the same time. Hence, we consider refitting the model without the penalty term for this given rank $r$:
\begin{align}
\label{eq:refit}
    \min_{U_i,D_i}  &\sum_{1\leq i<j\leq I} w_{ij} \|\hat{S}_{ij}-V_iD_iD_j^{\intercal}V_j^{\intercal}\|_F^2 \\
    &s.t. \quad V_i^{\intercal}V_i = I \quad \forall 1\leq i \leq I
\end{align}
Similarly, an iterative procedure can be applied. The subproblem for updating $V_i$ would be the same as \eqref{eq:U}. In terms of updating 
$D_i,i=1,\cdots,I$, the subproblem becomes 
\begin{equation*}  \min_{d_{ik},i=1,\cdots,I,k=1,\cdots,r}\sum_{k=1}^{r}\sum_{1\leq i<j \leq I}w_{ij}\left((\tilde{S}_{ij})_{kk}-d_{ik}d_{jk}\right)^2
\end{equation*}
which is similar to problem \eqref{eq:dk} and hence we can follow the same procedure \eqref{eq:dk}-\eqref{eq:d11} to obtain the solution for $d_{1k}$ as follows:

\begin{equation}
\label{eq:d1_refit}
\begin{split}
    \min_{d_{1k}}&\sum_{1<j \leq I}w_{1j}\left((\tilde{S}_{1j})_{kk}-d_{1k}d_{jk}\right)^2\\
    &s.t. \quad d_{1k}\geq 
0    \end{split}
\end{equation}

\begin{equation}
 d_{1k}=\begin{cases}
 \frac{\sum_{1<j\leq I}w_{1j}d_{jk}(\tilde{S}_{1j})_{kk}}{\sum_{1<j\leq I}w_{1j}d_{jk}^2} , & \sum_{1<j\leq I}w_{1j}d_{jk}(\tilde{S}_{1j})_{kk}>0 \quad \text{and} \quad \sum_{1<j\leq I}w_{1j}d_{jk}^2\neq 0\\
 0 &\sum_{1<j\leq I}w_{1j}d_{jk}(\tilde{S}_{1j})_{kk}<0 \quad \text{or} \quad\sum_{1<j\leq I}w_{1j}d_{jk}^2= 0
 \end{cases}
\end{equation}
$d_{jk},j=2,\cdots,k=1,\cdots,I$ can be obtained in the same way.

\section{Simulation Study} \label{sec:simulation}
In this section, we conduct simulation studies to evaluate the performance of joint\_LCA. We also apply JIVE with two options for rank selection, denoted as JIVE\_BIC and JIVE\_perm respectively. In addition, mCIA, \cite{min2020sparse} and multiple canonical correlation analysis (mCCA, \cite{witten2009extensions}) which is a direct extension of standard CCA to multiple data views, are implemented for comparison. 

For all settings, we generate $I$ data views $X_i \in \mathbb{R}^{n\times p_i}$ as follows:
\begin{equation}
\label{dat:gen}
\begin{split}
X_i &= Z_i + W_i, \\
Z_i &= U(V_iD_i)^{\intercal} + U_{i0}(V_{i0}D_{i0})^{\intercal},
\end{split}
\end{equation}
where two types of structures are encoded into the the true signal $Z_i$: a joint component ($U(V_iD_i)^{\intercal}$) and an individual component ($U_{i0}(V_{i0}D_{i0})^{\intercal}$). More specifically, $U$ corresponds to the common score matrix shared by $I$ data views, and $V_i$ is the view-specific loading matrix associated with the joint structure. All score matrices and loading matrices are generated using standard normal distribution with subsequent centering and orthonormalization. Each entry in the error matrix $W_i$ is generated from $\mathcal{N}(0, \sigma^2)$, where the noise level is set such that the overall signal to noise ratio equals to one:
\begin{equation*}
    \frac{\sum_{i=1}^I\|Z_i \|_F^2}{n(\sum_{i=1}^I p_i)\sigma^2}=1.
\end{equation*}
We denote the rank of the joint component and the rank of the individual structure associated with the $i$th data view as $r_0$ and $r_i$ respectively. In all simulation settings, the rank of all individual structures is set as $r_i=1, \forall i$.

In the simulation studies, we hope to investigate how different methods perform as the underlying rank of the joint structure $r_0$ varies. Specifically, we consider: 1) $r_0=2$: the rank of the joint structure is comparable to the rank of the individual structure; 2) $r_0 = 5$: the rank of the joint structure is relatively larger than that of the individual structure. Another aim is to study how the strength of the joint signal might affect the estimation results. To that end, we generate the diagonal matrices $D_i, 
D_{i0}$ in two ways. In case I, the diagonal entries for all diagonal matrices are generated from the the standard uniform distribution and hence the joint signal and the individual signal has comparable strength. In case II, the diagonal entries of the diagonal matrices associated with the joint structures ($D_i, 1\leq i \leq I$) and those of the diagonal matrices associated with the individual structures ($D_{i0}, 1\leq i \leq I$) are generated from different uniform distributions:
\begin{equation*}
    D_i \sim\mathcal{U}[0.5\sqrt{5}, \sqrt{5}], 
    D_{i0}\sim \mathcal{U}[0.5, 1], 1 \leq i \leq I,
\end{equation*}
where the percentage of variation explained by the joint structure is relatively larger than that explained by the individual structure. The sample size $n$ is another varying factor and we choose $n$ over $\left\{100, 200\right\}$.

To access how well joint\_LCA and JIVE perform in terms of determining the rank of joint component, we look into the rank selection accuracy, namely the proportion of replications in which the true rank is identified. In addition, we are interested in its performance in recovering the view-specific loading matrix $V_i$ in comparison to mCIA and mCCA. Specifically, we evaluate the scaled squared Frobenius norm error between the subspace spanned by true loadings $V_i$ and the subspace spanned by its estimate $\hat{V}_i$, defined as
\begin{equation}
\label{eq:est_err}
   \sum_{i=1}^I \frac{\|V_iV_i^{\intercal} - \hat{V}_i\hat{V}_i^{\intercal} \|_F^2}{I\|V_iV_i^{\intercal} \|_F^2}.
\end{equation}
JIVE gives signal matrix estimation associated with the joint structure for each data view instead of view-specific loading matrices. We perform SVD on the estimated joint signal matrix for each view and the first three right singular vectors are together used as an estimate for $V_i$.
As for sequential methods mCIA and mCCA which estimates one single loading at one time, for fair comparison, we extract $r_0$ components sequentially for each method, which are then concatenated horizontally to form the corresponding $\hat{V}_i$. 

When implementing joint\_LCA, the weight associated with each cross covariance error term is set as:
\begin{equation*}
    w_{ij} = \frac{1}{\| \hat{\Sigma}_{ij}\|_F^2}
\end{equation*}
where $\hat{\Sigma}_{ij}$ corresponds to the estimate of cross covariance between $X_i$ and $X_j$. Specifically for the tuning parameter selection, five-fold cross validation approach is employed to choose the best tuning parameter $\lambda$. In each simulation, we split the simulated multiview data into five folds. Each time the $k$th fold is held out as the testing set, and we obtain the estimate $\hat{V}_i$, $\hat{D}_i$ from the remaining samples, which are then used to construct the pairwise cross covariance estimate for the training data:
\begin{equation*}
    \hat{\Sigma}^{(-k)}_{ij} = \hat{V}_i\hat{D}_i\hat{D}_j^{\intercal}\hat{V}_j^{\intercal}.
\end{equation*}
The weighted sum of the squared Frobenius norm error between pairwise cross covariance estimate $\hat{\Sigma}^{(-k)}_{ij}$ from the training data and the standard cross covariance estimate $\hat{\Sigma}^{(k)}_{ij}$ from the testing data 
\begin{equation*}
    \sum_{1\leq i<j \leq I} w_{ij} \|\hat{\Sigma}^{(-k)}_{ij}-\hat{\Sigma}^{(k)}_{ij} \|_F^2
\end{equation*}
is used as the criteria for selecting optimal $\lambda$. In practice, we found that the tuning parameter derived this way usually leads to an overestimation of joint rank $r_0$.
In order to alleviate the overfitting issue, we apply one standard error rule to the cross validated estimate. More precisely, we select the largest tuning parameter within one standard error of the optimal tuning parameter yielded by the cross validation criteria.

\subsection{Three Data Views}
First, we consider $I=3$ data views. Apart from three design factors mentioned in the last section ($r_0$, $n$ and the way that $D_i$ and $D_{i0}$ are generated), we also consider two experimental designs when setting up the dimension for each data view: 1) balanced design: $p_1=p_2=p_3=100$; 2) unbalanced design: $p_1=100, p_2=200, p_3=300$. 

Results of rank selection accuracy for joint\_LCA, JIVE\_perm and JIVE\_BIC are shown in table \ref{tb_rs:3dat_r2} and \ref{tb_rs:3dat_r5}. It can been seen that the proposed joint\_LCA gives consistently good and robust rank estimation across all scenarios. In particular, joint\_LCA correctly identifies the rank of the joint signal in at least 90 percent replications in the setting where $r_0=2$. One observation that might be a little surprising is that the increase in sample size does not necessarily lead to improvement in rank selection accuracy for joint\_LCA. There might be two reasons underlying this phenomenon. First, an increase in the sample size will add more information to both the joint signal and the individual signal, which does not necessarily result in a larger separation between these two types of signals. Second, we adopt one standard error rule in the rank selection procedure to reduce overfitting and the standard error might get smaller as the sample size increases, making it harder to move the original optimal estimate selected by the cross validation criteria. JIVE\_perm and JIVE\_BIC also give comparably decent results in terms of rank selection in most scenarios, however, there are a few cases where either JIVE\_perm or JIVE\_BIC yields significantly worse rank estimation. JIVE\_perm tends to overestimate the rank of the joint structure in case I where the strength of the joint signal is comparable to that of the individual signal and an increase in the sample size $n$ or the total dimension of data views fails to improve its performance in the rank selection. While JIVE\_BIC performs very well when $n=200$, scenarios with a smaller sample size ($n=100$) and a larger $r_0$ seems to be more challenging to JIVE\_BIC, where it underestimates $r_0$ in the majority of replications.  

\begin{figure}[H]
\caption{Displayed are boxplots of estimation error $\sum_{i=1}^{3}\|\hat{V_i}\hat{V_i}^{\intercal}-V_iV_i^{\intercal} \|_2^2/3\|V_iV_i^{\intercal} \|_2^2$ for joint\_LCA, JIVE\_perm, JIVE\_BIC, mCCA and mCIA
where there are $I=3$ data views and the rank of the joint structure is $r_0=2$. Panel (a) and (b) show results for case I where $D_i$ and $D_{i0}$ are all generated from the standard uniform distribution; panel (c) and (d) show results for case II where $D_i$ and $D_{i0}$ are generated from the uniform distribution based on $[0.5\sqrt{5},\sqrt{5}]$ and $[0.5, 1]$ respectively.}
           \centering            \includegraphics[width=18cm,height=20.26cm]{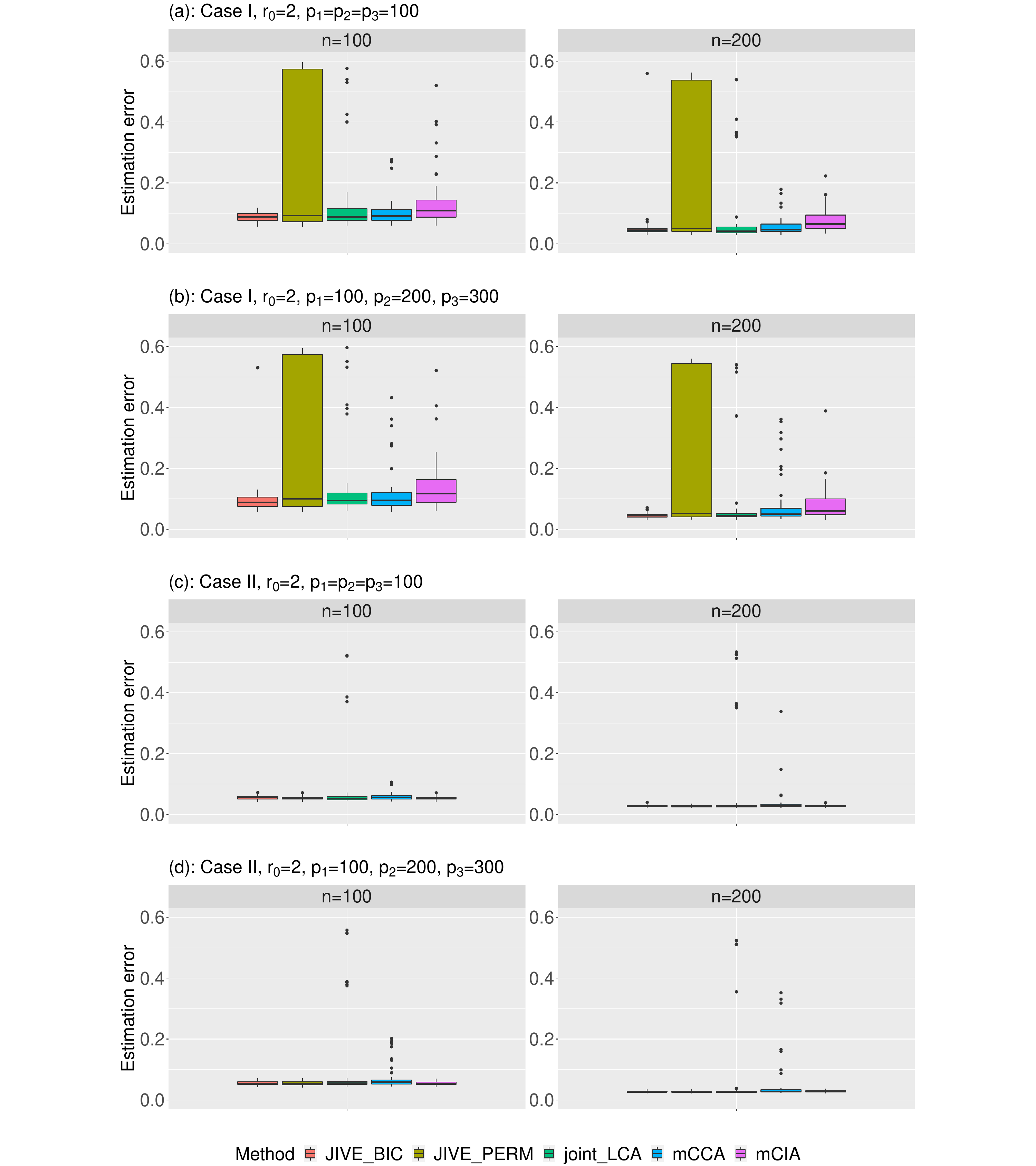}
            \label{fig_err:3dat_r2}
\end{figure}

\begin{figure}[H]
\caption{Displayed are boxplots of estimation error $\sum_{i=1}^{3}\|\hat{V_i}\hat{V_i}^{\intercal}-V_iV_i^{\intercal} \|_2^2/3\|V_iV_i^{\intercal} \|_2^2$ for joint\_LCA, JIVE\_perm, JIVE\_BIC, mCCA and mCIA
where there are $I=3$ data views and the rank of the joint structure is $r_0=5$. Panel (a) and (b) show results for case I where $D_i$ and $D_{i0}$ are all generated from the standard uniform distribution; panel (c) and (d) show results for case II where $D_i$ and $D_{i0}$ are generated from the uniform distribution based on $[0.5\sqrt{5},\sqrt{5}]$ and $[0.5, 1]$ respectively.}
           \centering            \includegraphics[width=18cm,height=20.26cm]{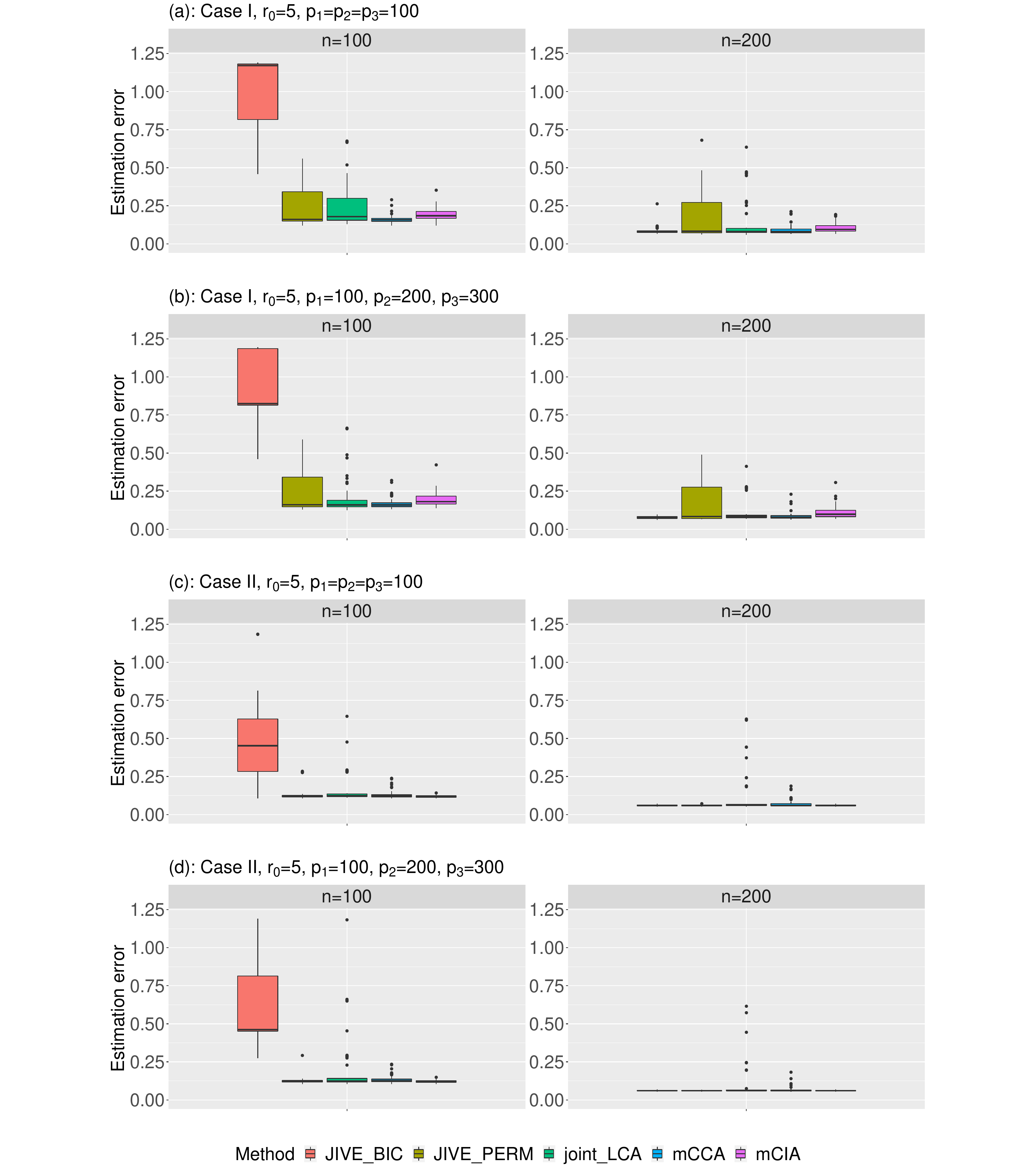}
            \label{fig_err:3dat_r5}
\end{figure}

Boxplots of the estimation error \eqref{eq:est_err} are displayed in figure \ref{fig_err:3dat_r2} and figure \ref{fig_err:3dat_r5}. On the first look, $r_0=5$ seems to be a more challenging case than $r_0=2$ for all methods with regard to the estimation of view-specific loading matrices, which is up to our expectation, since the number of parameters in the view-specific loading matrices would increase in the number of common components $r_0$. The estimates given by joint\_LCA yield consistently small errors with low variances under all combinations of $n$, $p$ and $r_0$ in case I and case II. Some outliers which are located outside the whiskers of the boxplot can be observed for joint\_LCA. The error outliers mostly come from those replications in which the rank of the joint signal is wrongly estimated, implying that the additional noisy columns incorrectly included in $\hat{V}_i$ might be responsible for these relatively larger estimation errors. The estimation given by JIVE\_BIC and JIVE\_perm is overall good in most settings. JIVE\_perm yields inaccurate estimates with larger variance in case I and JIVE\_BIC significantly underperforms all other methods when $r_0=5$ and $n=100$, which is as expected considering their unsatisfactory performance in terms of rank selection in these scenarios. As for mCCA and mCIA, even if the corresponding $\hat{V}_i$ is constructed given the true rank, there still exist a few outliers which are most likely attributed to the inefficiency in estimating the signal, suggesting that joint\_LCA is relatively more robust in the signal estimation compared to these two sequential methods. mCIA performs relatively worse than joint\_LCA and mCCA, especially in case I. The difference in the performance between mCIA and the other two methods might stem from the fact that joint\_LCA and mCCA are cross covariance based methods while the formulation of mCIA starts from individual data matrices. And the results suggest that the cross covariance based methods seem more preferable in terms of loading matrices estimation under our generative model \ref{dat:gen}.


\renewcommand{\arraystretch}{1.5}
\begin{table}[H]
 \centering
  \fontsize{12}{12}\selectfont
  \begin{threeparttable}
  \caption{Results of rank selection accuracy for joint\_LCA, JIVE\_perm and JIVE\_BIC over 100 replications in the scenarios where there are $I=3$ data views and the rank of joint structure $r_0=2$. Experimental factors include 1) sample size: $n\in \left\{100, 200\right\}$; 2) feature dimensions: $(p_1, p_2, p_3) \in \left\{\text{Balanced } (100, 100, 100), \text{Unbalanced } (100, 200, 300)\right\}$; 3) the way that the diagonal matrices $D_i$ and $D_{i0}$ are generated: case I and case II.
  }
 \label{tb_rs:3dat_r2}
    \begin{tabular}{ccccccc}
    \toprule
     & &
    \multicolumn{2}{c}{Case I }&\multicolumn{2}{c}{Case II}\\
    \cmidrule(lr){3-4} \cmidrule(lr){5-6} 
      Method & Sample size & Balanced  &Unbalanced&Balanced &Unbalanced\cr
    \midrule
\multirow{2}{*}{joint\_LCA}&$n=100$ &0.94 &0.92 &0.96 &0.94
    \\ &$n=200$ &0.92 &0.90 &0.90 &0.90\cr
\midrule 
 \multirow{2}{*}{JIVE\_perm}&$n=100$ &0.58 &0.56 &1.00 &1.00
    \\ &$n=200$ &0.50 &0.42 &1.00 &1.00 \cr
\midrule
    \multirow{2}{*}{JIVE\_BIC}&$n=100$ &1.00 &0.96 &1.00 &1.00
    \\ &$n=200$  &0.98 &1.00 &1.00 &1.00\cr
 \bottomrule
    \end{tabular}
    \end{threeparttable}
\end{table}

\renewcommand{\arraystretch}{1.5}
\begin{table}[H]
 \centering
  \fontsize{12}{12}\selectfont
  \begin{threeparttable}
  \caption{Results of rank selection accuracy for joint\_LCA, JIVE\_perm and JIVE\_BIC over 100 replications in the scenarios where there are $I=3$ data views and the rank of joint structure $r_0=5$. Experimental factors include 1) sample size: $n\in \left\{100, 200\right\}$; 2) feature dimensions: $(p_1, p_2, p_3) \in \left\{\text{Balanced } (100, 100, 100), \text{Unbalanced } (100, 200, 300)\right\}$; 3) the way that the diagonal matrices $D_i$ and $D_{i0}$ are generated: case I and case II.}
\label{tb_rs:3dat_r5}
    \begin{tabular}{ccccccc}
    \toprule
     & &
    \multicolumn{2}{c}{Case I }&\multicolumn{2}{c}{Case II}\\
    \cmidrule(lr){3-4} \cmidrule(lr){5-6} 
      Method & Sample size & Balanced  &Unbalanced&Balanced &Unbalanced\cr
    \midrule
\multirow{2}{*}{joint\_LCA}&$n=100$ &0.68 &0.80 &0.82 &0.78
    \\ &$n=200$ &0.86 &0.84 &0.88 &0.94\cr
\midrule 
 \multirow{2}{*}{JIVE\_perm}&$n=100$ &0.62 &0.62 &0.92 &0.98
    \\ &$n=200$ &0.62 &0.58 &1.00 &1.00 \cr
\midrule
    \multirow{2}{*}{JIVE\_BIC}&$n=100$ &0.00 &0.00 &0.14 &0.00
    \\ &$n=200$  &0.98 &1.00 &1.00 &1.00\cr
 \bottomrule
    \end{tabular}
    \end{threeparttable}
\end{table}

\subsection{Four Data Views}
We then look into a setting where the multiview data consists of $I=4$ sources. Similar to the scenario where we have three data views, we consider both balanced and unbalanced experiment design. Specifically, the dimensions for different data views are either set as equal $(p_1 = p_2 = p_3 = p_4 = 100)$ or unequal $(p_1 = 100, p_2 = 200, p_3 = 300, p_4 = 400)$.

\begin{figure}[H]
\caption{Displayed are boxplots of estimation error $\sum_{i=1}^{3}\|\hat{V_i}\hat{V_i}^{\intercal}-V_iV_i^{\intercal} \|_2^2/4\|V_iV_i^{\intercal} \|_2^2$ for joint\_LCA, JIVE\_perm, JIVE\_BIC, mCCA and mCIA
where there are $I=4$ data views and the rank of the joint structure is $r_0=2$. Panel (a) and (b) show results for case I where $D_i$ and $D_{i0}$ are all generated from the standard uniform distribution; panel (c) and (d) show results for case II where $D_i$ and $D_{i0}$ are generated from the uniform distribution based on $[0.5\sqrt{5},\sqrt{5}]$ and $[0.5, 1]$ respectively.}
           \centering            \includegraphics[width=18cm,height=20.26cm]{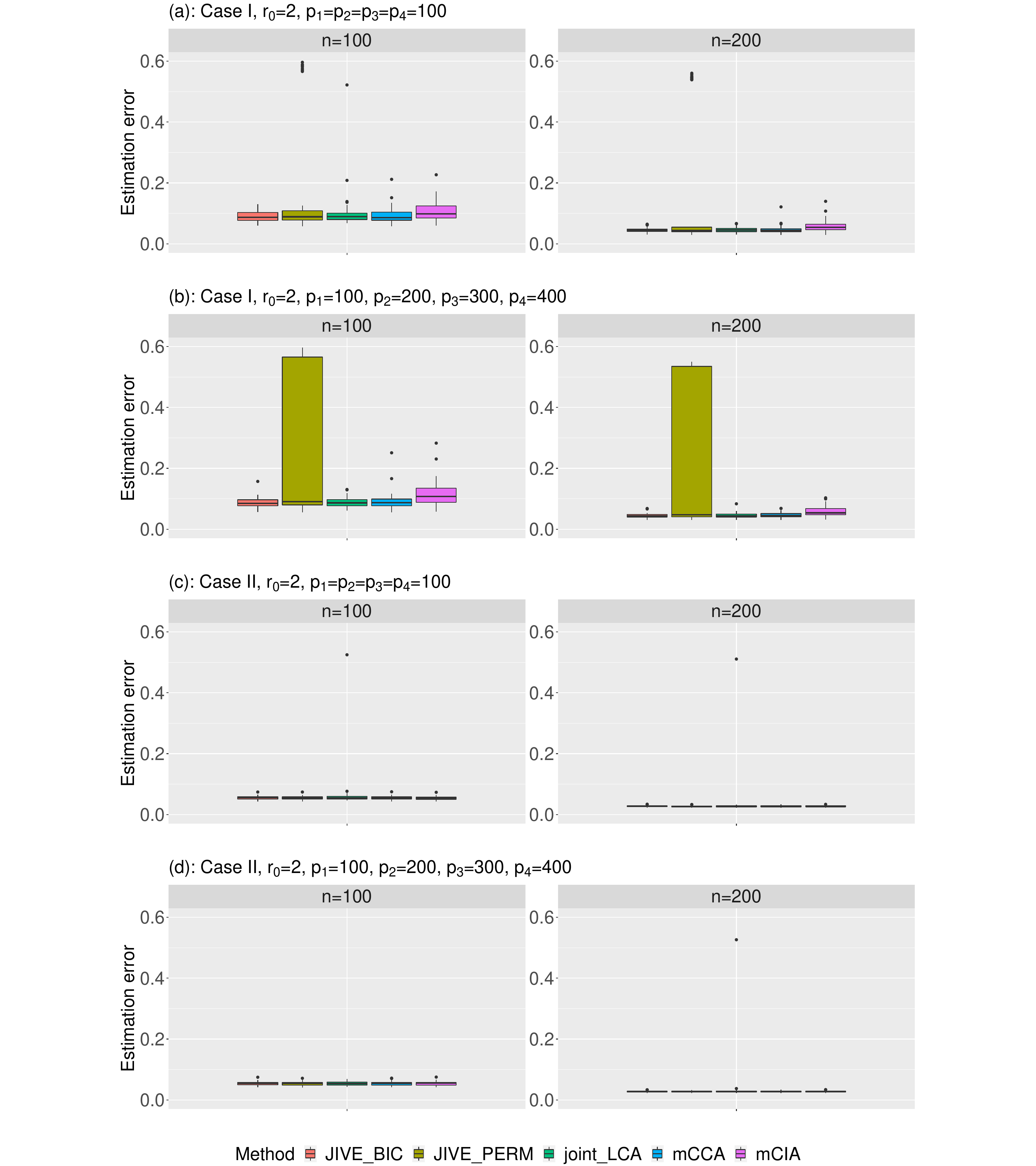}
            \label{fig_err:4dat_r2}
\end{figure}

\begin{figure}[H]
\caption{Displayed are boxplots of estimation error $\sum_{i=1}^{3}\|\hat{V_i}\hat{V_i}^{\intercal}-V_iV_i^{\intercal} \|_2^2/4\|V_iV_i^{\intercal} \|_2^2$ for joint\_LCA, JIVE\_perm, JIVE\_BIC, mCCA and mCIA
where there are $I=3$ data views and the rank of the joint structure is $r_0=5$. Panel (a) and (b) show results for case I where $D_i$ and $D_{i0}$ are all generated from the standard uniform distribution; panel (c) and (d) show results for case II where $D_i$ and $D_{i0}$ are generated from the uniform distribution based on $[0.5\sqrt{5},\sqrt{5}]$ and $[0.5, 1]$ respectively.}
           \centering            \includegraphics[width=18cm,height=20.26cm]{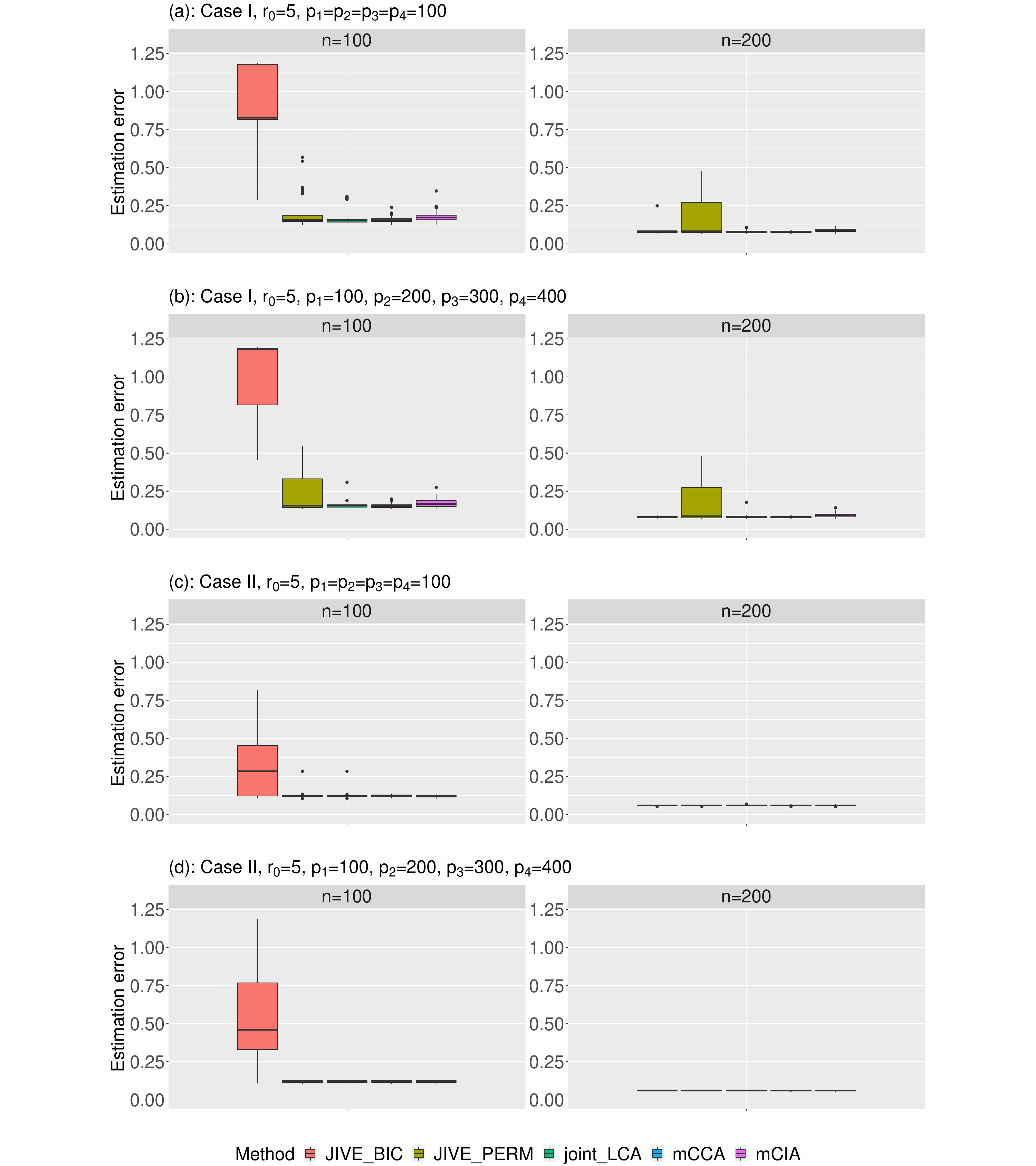}
            \label{fig_err:4dat_r5}
\end{figure}

Results in table \ref{tb_rs:4dat_r2} and table \ref{tb_rs:4dat_r5} show that the joint\_LCA correctly identifies the rank for the joint structure in over 90\% replications across all scenarios. Moreover, a boost in the estimation performance can also be observed in figure \eqref{fig_err:4dat_r2} and figure \eqref{fig_err:4dat_r5}: additional information from another data view help yield more accurate (smaller median estimation error) and more stable (lower estimation variance) estimation of the view-specific loading matrices for all methods. Specifically for joint\_LCA, there exist fewer and less extreme outliers mainly due to an enhancement in the rank selection accuracy. These observations illustrate the power of integrating more related data sources into analysis. As for JIVE\_PERM and JIVE\_BIC, while there are slight improvements in the scenarios where they fail to do well when only three data views are in use, their performance is still much inferior to that of joint\_LCA in terms of both rank identification and signal estimation.

\renewcommand{\arraystretch}{1.5}
\begin{table}[H]
 \centering
  \fontsize{12}{12}\selectfont
  \begin{threeparttable}
  \caption{Results of rank selection accuracy for joint\_LCA, JIVE\_perm and JIVE\_BIC over 100 replications in the scenarios where there are $I=4$ data views and the rank of joint structure $r_0=2$. Experimental factors include 1) sample size: $n\in \left\{100, 200\right\}$; 2) feature dimensions: $(p_1, p_2, p_3, p_4) \in \left\{\text{Balanced } (100, 100, 100, 100), \text{Unbalanced } (100, 200, 300, 400)\right\}$; 3) the way that the diagonal matrices $D_i$ and $D_{i0}$ are generated: case I and case II.}
 \label{tb_rs:4dat_r2}
    \begin{tabular}{ccccccc}
    \toprule
     & &
    \multicolumn{2}{c}{Case I }&\multicolumn{2}{c}{Case II}\\
    \cmidrule(lr){3-4} \cmidrule(lr){5-6} 
      Method & Sample size & Balanced  &Unbalanced&Balanced &Unbalanced\cr
    \midrule
\multirow{2}{*}{joint\_LCA}&$n=100$ &0.98 &1.00 &0.98 &1.00
    \\ &$n=200$ &1.00 &1.00 &0.98 &0.98\cr
\midrule 
 \multirow{2}{*}{JIVE\_perm}&$n=100$ &0.72 &0.58 &1.00 &1.00
    \\ &$n=200$ &0.68 &0.62 &1.00 &1.00 \cr
\midrule
    \multirow{2}{*}{JIVE\_BIC}&$n=100$ &1.00 &1.00 &1.00 &1.00
    \\ &$n=200$  &1.00 &1.00 &1.00 &1.00\cr
 \bottomrule
    \end{tabular}
    \end{threeparttable}
\end{table}

\renewcommand{\arraystretch}{1.5}
\begin{table}[H]
 \centering
  \fontsize{12}{12}\selectfont
  \begin{threeparttable}
  \caption{Results of rank selection accuracy for joint\_LCA, JIVE\_perm and JIVE\_BIC over 100 replications in the scenarios where there are $I=4$ data views and the rank of joint structure $r_0=5$. Experimental factors include 1) sample size: $n\in \left\{100, 200\right\}$; 2) feature dimensions: $(p_1, p_2, p_3, p_4) \in \left\{\text{Balanced } (100, 100, 100, 100), \text{Unbalanced } (100, 200, 300, 400)\right\}$; 3) the way that the diagonal matrices $D_i$ and $D_{i0}$ are generated: case I and case II.}
 \label{tb_rs:4dat_r5}
    \begin{tabular}{ccccccc}
    \toprule
     & &
    \multicolumn{2}{c}{Case I }&\multicolumn{2}{c}{Case II}\\
    \cmidrule(lr){3-4} \cmidrule(lr){5-6} 
      Method & Sample size & Balanced  &Unbalanced&Balanced &Unbalanced\cr
    \midrule
\multirow{2}{*}{joint\_LCA}&$n=100$ &0.92 &0.98 &0.98 &1.00
    \\ &$n=200$ &1.00 &1.00 &1.00 &1.00\cr
\midrule 
 \multirow{2}{*}{JIVE\_perm}&$n=100$ &0.80 &0.70 &1.00 &1.00
    \\ &$n=200$ &0.60 &0.64 &1.00  &1.00\cr
\midrule
    \multirow{2}{*}{JIVE\_BIC}&$n=100$ &0.00 &0.00 &0.32 &0.02
    \\ &$n=200$  &0.98 &1.00 &1.00 &1.00\cr
 \bottomrule
    \end{tabular}
    \end{threeparttable}
\end{table}

\section{Real Data Application}\label{sec:real_data}
\subsection{Nutrimouse Data}
The nutrimouse data set comes from a nutrition study in mouse (\cite{martin2007novel}). In this study, two sets of variables were acquired from forty mice.
In specific, one feature set contains expressions of 120 cells measured in liver cells, and the other set concerns measurements of concentrations of 21 hepatic fatty acids (FA). In addition, biological units (mice) are cross-classified according to two factors:
\begin{itemize}
    \item Genotype: wide-type (WT) mice and PPAR$\alpha$ deficient (PPAR$\alpha$) mice
    \item Diet: reference diet (REF), saturated FA diet (COC), $\omega 6$ FA-rich diet (SUN), $\omega 3$ FA-rich diet (LIN), and the FISH diet.
\end{itemize}
The Nutrimouse data view is available within package CCA.
\begin{figure}[hbt!]%
    \centering
\subfloat{{\includegraphics[width=8cm, height=8cm]{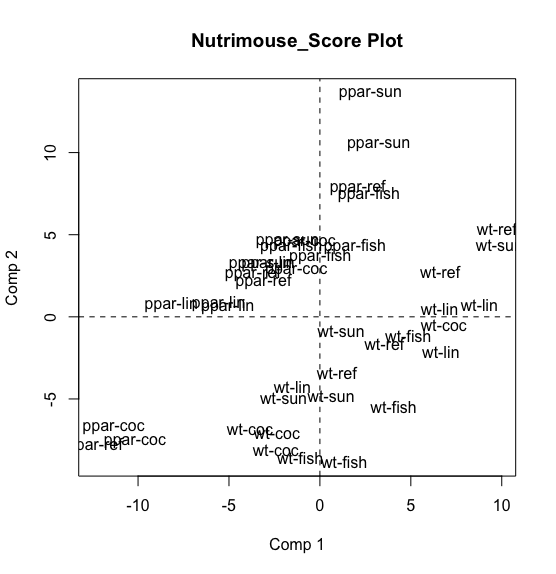} }}%
    \subfloat{{\includegraphics[width=8cm, height=8cm]{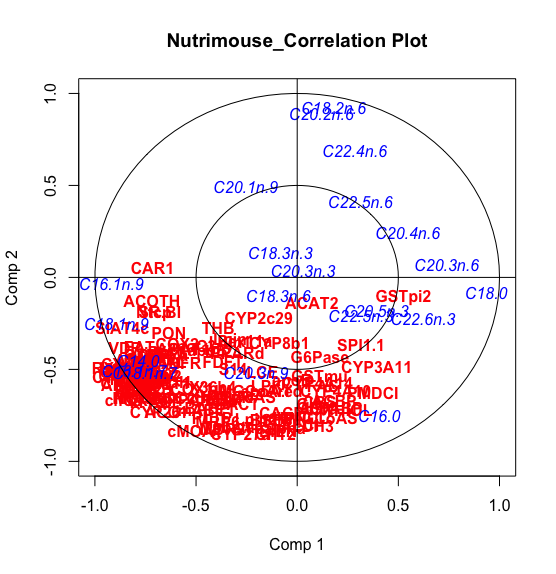} }}%
    \caption{Nutrimouse}%
    \label{nutri}
\end{figure}
We applied our method to the Nutrimouse data, and the rank of the common space is estimated to be two. We project the each data matrix onto these two dimensions and we also plot the correlation between the projected score and the original features to see the contribution from each feature to each common component, which is shown in figure \eqref{nutri}. The left panel shows a clear separation of genotypes along the second dimension, though, there are a few overlaps. This agrees with the finding in \cite{martin2007novel} that PPAR$\alpha$ mice have high concentrations of linoleic acid (C18.2n.6), which is also consistent with the observation in the right panel that C18.2n.6 has high positive coordinates on the second dimension.

\subsection{Boston Housing Data}
Boston Housing data set (\url{http://math.furman.edu/~dcs/courses/math47/R/library/mlbench/html/BostonHousing.html}) which contains $N=506$ 14-dimensional instances. The predictors can be divided into 3 different data sets with dimensions $n_1 = 4, n_2 =6, n_3 =3$:

\begin{itemize}
    \item $X_1$ (AN, AGE, TAX, RM): Variables directly related with the housing market.
    \item $X_2$ (CRIM, INDUS, NOX, PRTATIO, B, LSTAT): Variables indirectly related with the housing the housing market.
    \item $X_3$ (CHAS, DIS, RAD): Geographical variables.
\end{itemize}
In addition, the target variable is MEDV, the median value of owner-occupied homes in USD. A detailed description of variables listed above can be found at \url{http://math.furman.edu/~dcs/courses/math47/R/library/mlbench/html/BostonHousing.html}

\begin{figure}[hbt!]%
    \centering
    {\includegraphics[width=8cm, height=8cm]{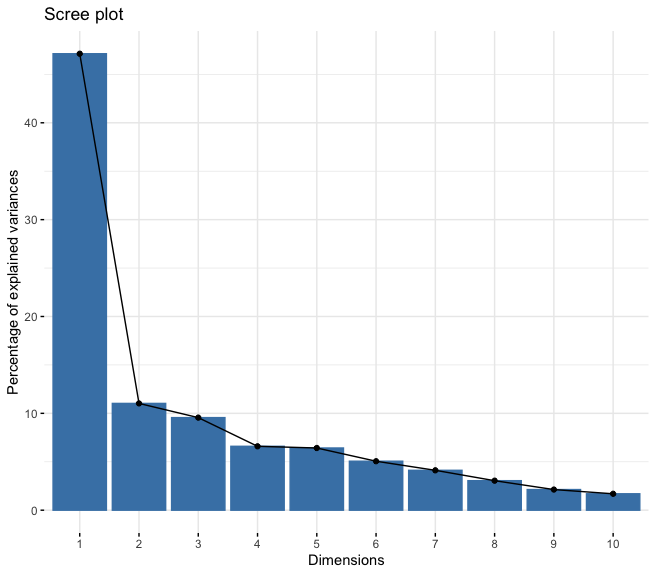} }%
    \caption{Scree Plot}%
    \label{pca_scree}
\end{figure}

This data view has been widely adopted to illustrate how to build a model to predict the housing values using the given features by different techniques.
We propose to construct a new feature set based on the score matrix from canonical correlation analysis and then use regression algorithms to predict the housing prices. Specifically, our method can be applied to the Boston Housing data and obtain the canonical loading matrices $\left\{V_1,V_2,V_3\right\}$ for $\left\{X_1,X_2,X_3\right\}$. The score matrix is calculated as $U=\frac{X_1V_1D_1^{-1}+X_2V_2D_2^{-1}+X_3V_3D_3^{-1}}{3}$, which is then fed into regression algorithms like support vector regression (SVR) and regression tree (CART). For the purpose of comparison, we also consider performing principle component analysis on the original feature sets first and using thee obtained principle components as the features. According to the scree plot \eqref{pca_scree}, we use either the first two/three principal components as the new feature set.

We compared the performance/effectiveness between different feature sets by the squared root of the mean squared error obtained from 10-fold cross validation. For the canonical variate features, the rMSE are 6.88/6.65 for the SVR model and CART respectively. 
For the principle component features, the rMSE corresponding to the SVR are 6.81 and 6.39 for the CART. The principal component features perform slightly better since it contains comprehensive information for all kinds of structures while canonical variate features are constructed based only on the common structure. 

\subsection{Russett Data}
The Russett data view contains three data views for 47 countries. This data view was collected in order to study the relationship between Agricultural Inequality, Industrial Development, and Political Instability.
\begin{itemize}
    \item $X_1$ (Gini, Farm, Rent): Variables related to industrial development
    \item $X_2$ (Gnpr, Labo): Variables that measure industrial development
    \item $X_3$ (Inst, Ecks, Deat, Demo): Variables that describe political instability. In particular, Demo is a categorical variable that describes the political regime: stable democracy, unstable democracy and dictatorship.
\end{itemize}
Two pairs of canonical variates are identified by our method. The left graphical plot is obtained by plotting the first pair of canonical variates ($X_1V_1$, $X_2V_2$), labeled by their political regime in 1960. Clearly, the first component exhibits a separation among regimes. For instance, the countries with dictatorships are concentrated in the upper right quadrant. It is worth noting that the factors labor and gini have a large positive contribution to the first dimension from the right plot. High percentage of labor force employed in agriculture indicates below average industrial development and high gini index is related to unequal land distribution. It is difficult for a country to escape dictatorship when its industrial development is low while its agricultural inequality is high. Our result also agrees with the finding in \cite{} some countries with unstable democracy labelled by green, (Greece, Brazil, Chile, Argentina ) locate in the first quadrant, which became dictatorships for a period of time after 1960.

\begin{figure}[hbt!]%
    \centering
    \subfloat{{\includegraphics[width=8cm, height=8cm]{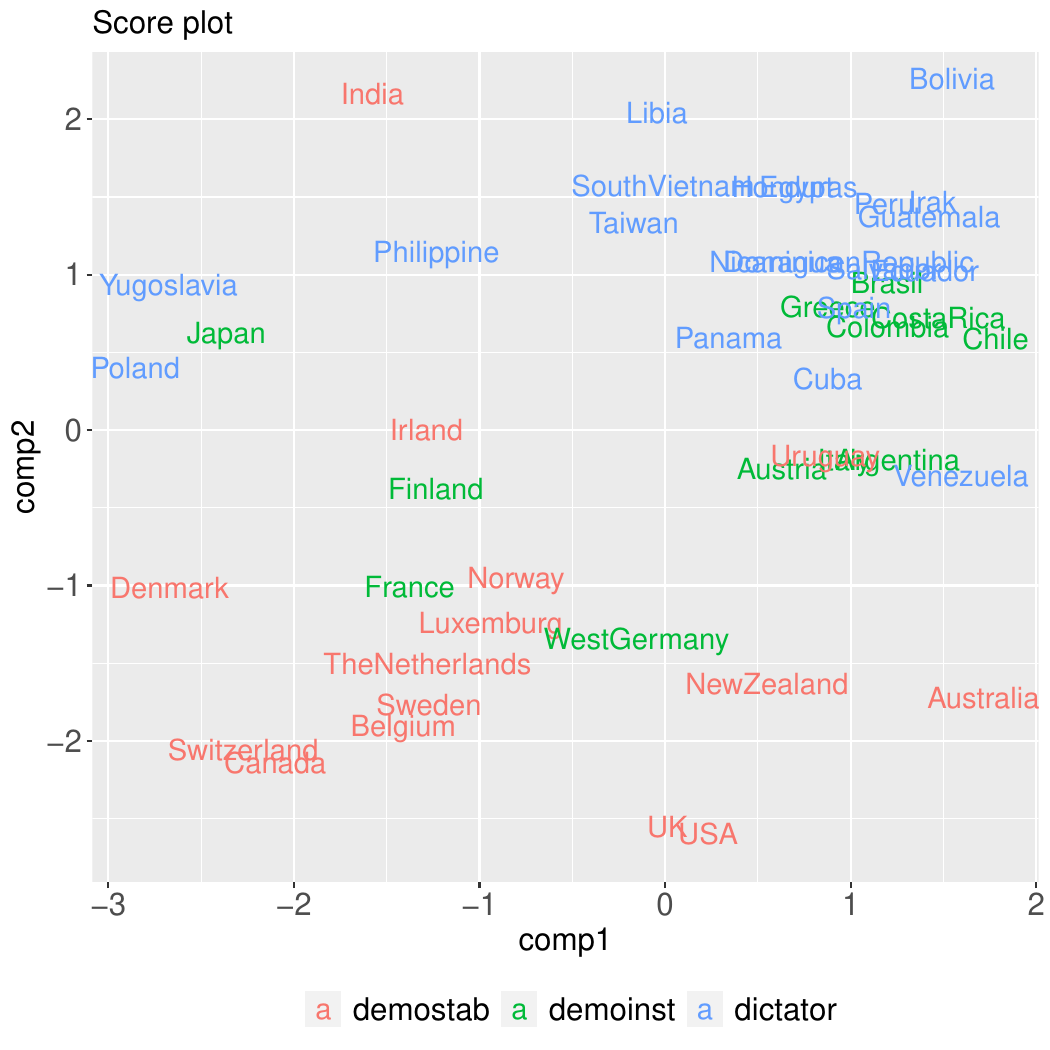} }}
     \subfloat{{\includegraphics[width=8cm, height=8cm]{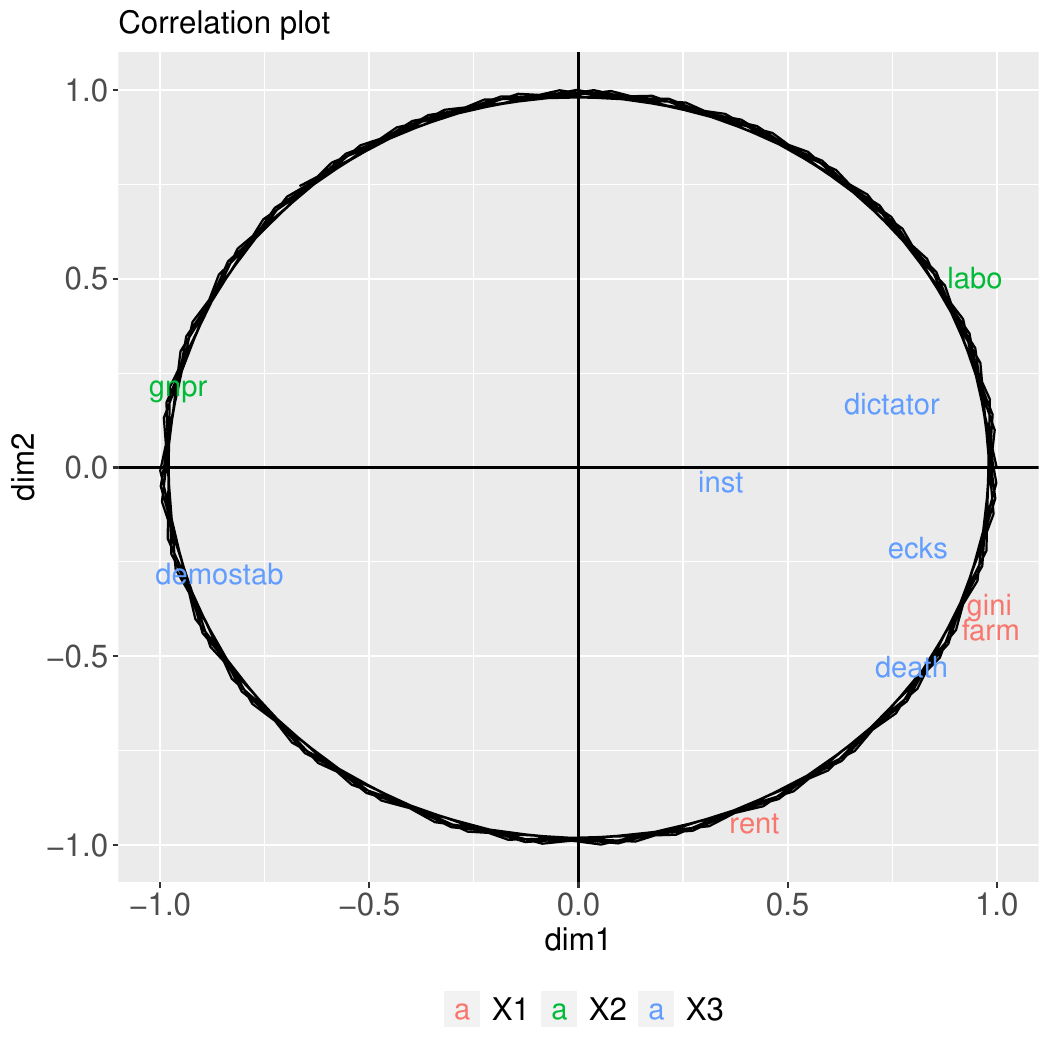} }}%
    \caption{Russett Data}%
    \label{Russett}
\end{figure}

\subsection{Multiview Single Cell Data}

\begin{figure}[H]
\caption{Scatterplot visualization of cells colored by cell types based on the first two directions obtained by tSNE.}
           \centering        \includegraphics[width=9cm,height=9cm]{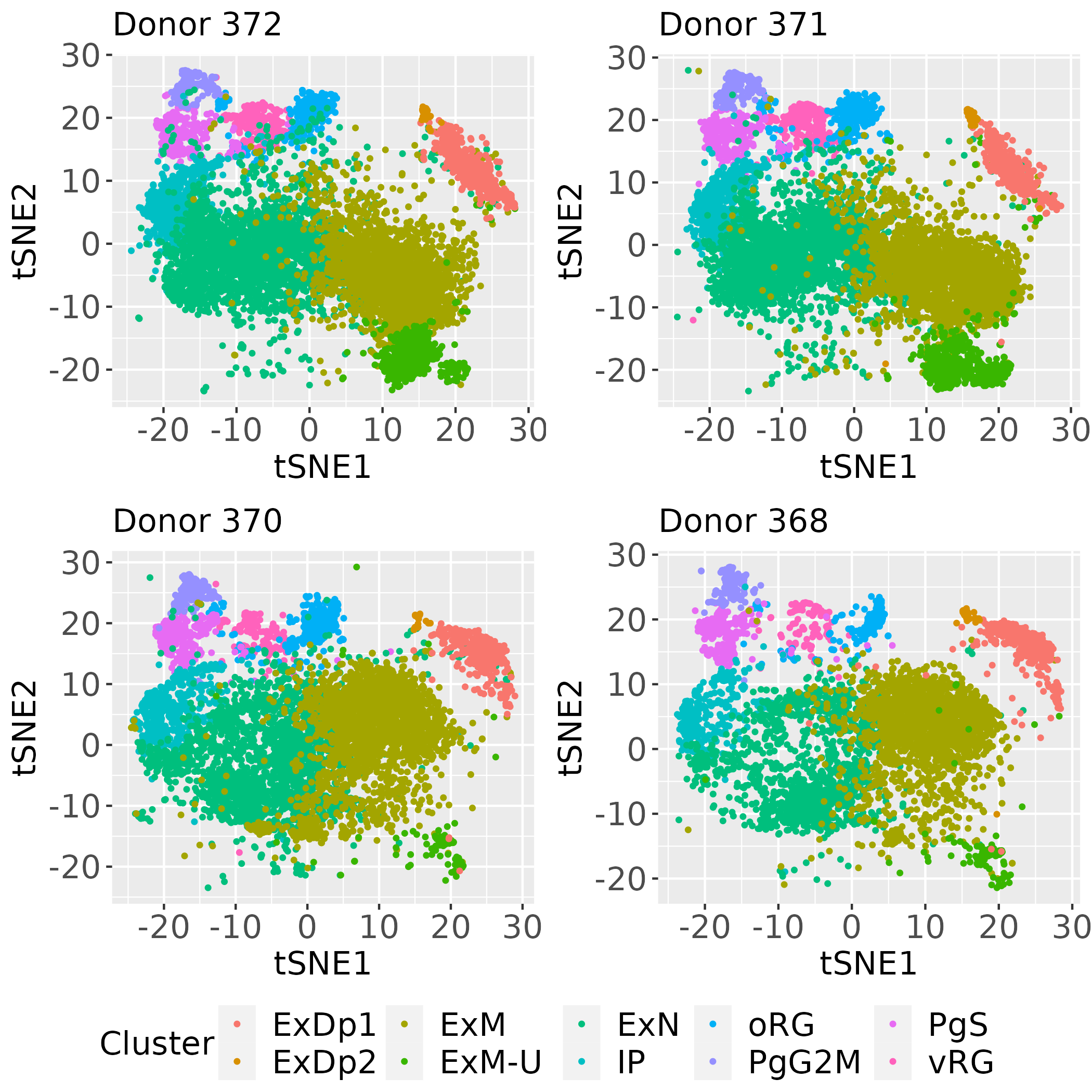}
            \label{fig:tsne_cluster}
\end{figure}

Next, we consider a collection of four single cell datasets from \cite{polioudakis2019single}. In this study, developing brain tissue samples were obtained from four patient donors (donor 372, donor 371, donor 370, donor 368). Dissection were first performed on fetal brain tissues and single cells were isolated for analysis. In this process, sixteen types of cells at different stages of neuronal differentiation and maturation were identified. Drop-seq \citep{macosko2015highly} were then run on single cells to obtain high quality gene expression profiles for around 40, 000 cells. This study provides us with the ability to deepen our understanding in human neurogenesis, cortical evolution and the cellular basis of neuropsychiatric disease \citep{polioudakis2019single}. Here, we only include ten major cell types for analysis. As a result, we have four data views:
\begin{equation*}
  X_1\in \mathbb{R}^{35543\times 8530}, X_2\in \mathbb{R}^{35543\times 9082}, X_3\in \mathbb{R}^{35543\times 7066}, X_4\in \mathbb{R}^{35543\times 5454}.
\end{equation*}
To illustrate the distribution of cell types, t-distributed stochastic neighbor embedding (tSNE) is applied to perform dimension reduction for each data view. Then we project each data view onto the two-dimensional subspace found by tSNE where each cell is colored by the corresponding cell type. As can be seen from Figure \ref{fig:tsne_cluster}, the cell types can be very well spatially segregated in this way.

\begin{figure}[H]
\caption{Scatterplot of cells colored by the corresponding value in the common component (a specific column in $\hat{V}_i\hat{D}_i, 1\leq i \leq 4$). Results for component 2, 5 and 7 are shown below.}
           \centering            \includegraphics[width=16cm,height=12cm]{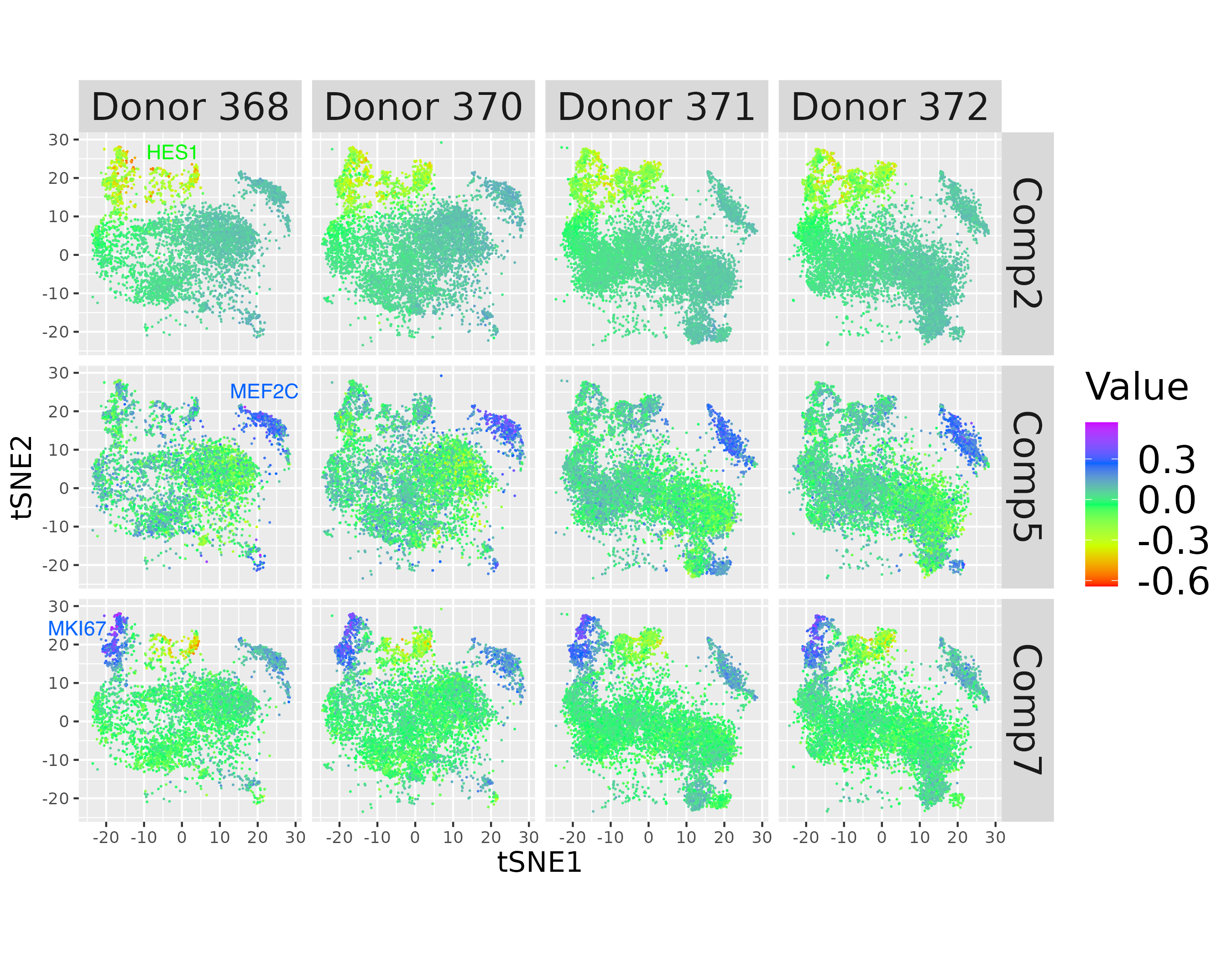}
           \label{fig:common}
\end{figure}

We apply joint\_LCA to this multiview single cell data. The estimated number of common components is $\hat{r}_0=8$ and we obtain view-specific estimation $\hat{V}_i$ and $\hat{D}_i$. We look into each column in the loading matrix multiplied by the diagonal matrix, i.e., $\hat{V}_i\hat{D}_i$, which is referred to as a common component. Based on the 2D scatterplot in Figure \ref{fig:tsne_cluster}, each cell point in Figure \ref{fig:common} is colored by the corresponding value in the common component. With respect to each row, the color distribution is quite similar across different data views, suggesting that the common components found capture shared information across different donors. In addition, we identify some important genes by looking at the correlation between the gene expression data $X_i$ and the common components $\hat{V}_i\hat{D}_i$, including genes HES1, MEF2C, MKI67.
For instance, gene MKI67 is most positively correlated with the 7th component. These three genes identified have been shown to be informative in the neurogenesis process (\cite{polioudakis2019single}; \cite{mathews2017evidence}). Additionally, we observe some clusters of cells with either the highest or the lowest values for each component in Figure \ref{fig:tsne_cluster}, which turn out to be connected with the identified genes. Specifically, for each common component, the identified gene tends to have high expression level in the clusters of cells which have the highest or lowest values in the component.

 \renewcommand{\arraystretch}{2}
\begin{table}[H]
 \centering
  \fontsize{10}{10}\selectfont
  \begin{threeparttable}
\caption{Mean expression level of genes HES1, MEF2C, MKI67 in each type of cells. }
\label{tb:gene_expression}
   \begin{tabular}{cccccccccccc}
   \toprule

          HES1  & \textbf{vRG}  &\textbf{oRG}   &PgS  &PgG2M  &IP  &ExN   &ExM  &ExM-U &ExDp1 &ExDp2
            \\ 
     
    &\textbf{7.32} &\textbf{7.71} &3.35 &2.79 &0.79 &0.16 &0.19 &0.13
     &0.13 &0.00    \\ 

\midrule
            MEF2C  & vRG  &oRG   &PgS  &PgG2M  &IP  &ExN   &ExM  &\textbf{ExM-U} &\textbf{ExDp1} &\textbf{ExDp2}
            \\ 
    
         &0.41  &0.35  &0.29  &0.29  &0.29  &0.48  &5.42
  &\textbf{9.24} &\textbf{10.27} &\textbf{10.10}   \\ 
    \midrule
           MKI67 & vRG  &oRG   &\textbf{PgS}  &\textbf{PgG2M}  &IP  &ExN   &ExM  &ExM-U &ExDp1 &ExDp2
            \\ 
   
       &1.68  &2.38  &\textbf{8.54} &\textbf{12.21} &1.47  &0.36  &0.15
 &0.23  &0.07  &0.17  \\  
 \bottomrule
\end{tabular}
 \end{threeparttable}
 \end{table}

\section{Discussion}\label{sec:discussion}
There has been growing interests and demands in studying heterogeneous data from multiple sources in a collective way. It is of particular interest to capture variation common to all data views, which is helpful to understand the scientific association between different data views. Existing CCA-based methods focus on projecting each data view onto a common latent subspace. Specifically, they take advantage of the pairwise cross covariance matrices, and the view-specific orthonormal loadings corresponding to the joint structure are extracted in a successive way. One drawback associated with these sequential techniques is lack of an objective way to determine the number of shared components, namely, the rank of the common latent subspace. We propose the joint linked component analysis (joint\_LCA) for multiview data, which seeks for a set of canonical variates for each data view while simultaneously determining the rank of the common latent subspace. A novel nuclear-norm based penalty is designed to perform rank selection and a simple refitting procedure is adopted to correct the shrinkage bias in our estimate. We investigate the empirical performance of joint\_LCA in various simulation settings including small and large number of common components, low and high strength of joint signal and different number of data views. In comparison to sequential CCA-based methods (mCCA, mCIA), or JIVE with two  schemes for rank selection, joint\_LCA yields consistently and reasonably well results across all scenarios in terms of rank selection and loading matrix estimation, whereas its competitors perform poorly in some of these settings. We also apply joint\_LCA to several real multiview data views to illustrate the power of joint\_LCA as a useful tool for exploring common relationships among multiple data views.

The proposed method can be generalized to high dimensional data sets where sparsity is desired. One straightforward way is to impose structural penalty on loading matrices $V_i$ in the objective function \eqref{eq:model1}. However, we expect this to be a very challenging problem from a computational perspective, due to the complex interaction between the sparsity penalty and the orthogonality constraint. \cite{kallus2019mm} deals with a similar problem. They address this challenge by representing each orthonormal matrix $V_i$ as a multiplication of Givens rotation matrices \citep{shepard2015representation} and the optimization of the new objective function is performed with Broyden–Fletcher–Goldfarb–Shanno (BFGS) algorithm. While the reparametrization can successfully reduce the number of parameters and remove the orthogonal constraints, the computational cost of this strategy is still too high to be scaled to large data sets. Hence, future work is needed to develop a computationally efficient method that extends \eqref{eq:model1} to accommodate high dimensional data.

\section{Appendix}

In this section, we want to establish the connection between our proposed method and the generalized canonical correlation methods in the literature. Our finding is that, in the case $I=2$, $r_0=1$, the objective function below:
\begin{equation*}
\label{eq:U}
    \begin{split}
    \min_{V_1,V_2,D_1,D_2}  \|\hat{S}_{12}-V_1D_1D_2^{\intercal}V_2^{\intercal}\|_F^2 \\
    \text{s.t.} \quad V_i^{\intercal}V_i = \boldsymbol{I}, i=1,2
    \end{split}
\end{equation*}
is equivalent to the formulation of diagonalized CCA proposed in \cite{witten2009penalized}
\begin{equation}
\label{eq:res}
\begin{split}
   \max_{V_1,\cdots,V_I} \sum_{i<j} V_i^{\intercal}X_i^{\intercal}X_jV_j \\ \text{s.t} \quad\| V_i\|^2\leq 1
   \end{split}
\end{equation}

If $I=3$ and $r_0=1$, the optimization problem 
\begin{equation*}
\label{eq:U}
    \begin{split}
    \min_{V_i,D_i,i=1,2,3}  \sum_{1\leq i<j\leq 3}\|\hat{S}_{ij}-V_iD_iD_j^{\intercal}V_j^{\intercal}\|_F^2 \\
    \text{s.t.} \quad V_i^{\intercal}V_i = \boldsymbol{I}, i=1,2,3
    \end{split}
\end{equation*}
is equivalent to the SSQCOV-1 problem in \cite{hanafi2006analysis}, which is defined as 
\begin{equation*}
\begin{split}
    \max_{V_1,V_2,V_3}\sum_{1\leq i<j \leq 3}\text{cov}^2(X_iV_i,X_jV_j)\\
  \text{s.t.} \quad \|V_i \|_2=1, i=1,2,3
  \end{split}  
\end{equation*}

Denote the objective function as $f=$
Suppose that the true rank is $r=1$,then $D_1=d_1$, $D_2=d_2$ then minimizing the objective function is equivalent to minimizing 
\begin{equation}
\label{eq:langr}
\begin{split}
f(V_1,V_2,D_1,D_2) = -2d_1d_2\langle\hat{S}_{12},V_1V_2^{\intercal} \rangle + d_1^2d_2^2\\
\text{s.t.} \quad -d_i\leq 0, V_i^{\intercal}V_i = \boldsymbol{I}, i=1,2
\end{split}
\end{equation}
Keeping $V_i,i=1,2$ fixed, the optimization problem with respect to $d_1$ and $d_2$ is 
\begin{equation*}
\begin{split}
f(D_1,D_2) = -2d_1d_2\langle\hat{S}_{12},V_1V_2^{\intercal} \rangle + d_1^2d_2^2\\
\text{s.t.} \quad -d_i\leq 0, i=1,2
\end{split}
\end{equation*}
The Langrangian multiplier associated with this problem is,
\begin{equation*}
    L(D_1,D_2,v_1,v_2)=-2d_1d_2\langle\hat{S}_{12},V_1V_2^{\intercal} \rangle + d_1^2d_2^2-v_1d_1-v_2d_2
\end{equation*}
The corresponding KKT conditions are:
\begin{equation*}
\begin{split}
 \frac{\partial L}{\partial d_1} = -2d_2\langle \hat{S}_{12},V_1V_2^{\intercal}\rangle+2d_1d_2^2-v_1=0\\
 \frac{\partial L}{\partial d_2} = -2d_1\langle \hat{S}_{12},V_1V_2^{\intercal}\rangle+2d_1^2d_2-v_2=0\\
 -v_id_i=0,i=1,2\\
 -d_i\leq 0, i=1,2
 \end{split}
\end{equation*}

\begin{equation*}
\begin{split}
d_1\frac{\partial L}{\partial d_1}&=-2d_1d_2\langle \hat{S}_{12},V_1V_2^{\intercal}\rangle+2d_1^2d_2^2-d_1v_1\\
&=-2d_1d_2\langle \hat{S}_{12},V_1V_2^{\intercal}\rangle+2d_1^2d_2^2\\
&=2d_1d_2[d_1d_2-\langle \hat{S}_{12},V_1V_2^{\intercal}\rangle]\\
&=0
\end{split}
\end{equation*}

Since $d_1>0,d_2>0$, we have 
\begin{equation}
\label{eq:nonneg}
    d_1d_2=\langle \hat{S}_{12},V_1V_2^{\intercal}\rangle>0
\end{equation}

Substituting this into \eqref{eq:langr}, 
\begin{equation*}
    f(V_1,V_2) = -(\langle \hat{S}_{12},V_1V_2^{\intercal}\rangle)^2
\end{equation*}
Due to \eqref{eq:nonneg}, this is equivalent to minimizing 
\begin{equation*}
    -\langle \hat{S}_{12},V_1V_2^{\intercal} \rangle = -V_1^{\intercal}\hat{S}_{12}V_2
\end{equation*}

Now suppose that we have three data views, the objective function becomes
\begin{equation*}
\label{eq:langr2}
\begin{split}
f(V_1,V_2,V_3,D_1,D_2,D_3) &= -2d_1d_2\langle\hat{S}_{12},V_1V_2^{\intercal} \rangle + d_1^2d_2^2\\
&-2d_1d_3\langle\hat{S}_{13},V_1V_3^{\intercal} \rangle + d_1^2d_3^2-2d_2d_3\langle\hat{S}_{23},V_2V_3^{\intercal} \rangle + d_2^2d_3^2\\
\text{s.t.} \quad -d_i\leq 0, V_i^{\intercal}V_i &= \boldsymbol{I}, i=1,2,3
\end{split}
\end{equation*}

Similarly, we obtain that
\begin{equation*}
    \begin{split}
\frac{\partial f}{\partial d_1} = -2d_2\langle \hat{S}_{12},V_1V_2^{\intercal}    \rangle+2d_1d_2^2-2d_3\langle \hat{S}_{13},V_1V_3^{\intercal} \rangle+2d_1d_3^2+v_1=0\\
\frac{\partial f}{\partial d_2} = -2d_1\langle\hat{S}_{12},V_1V_2^{\intercal}    \rangle+2d_1^2d_2-2d_3\langle \hat{S}_{23},V_2V_3^{\intercal} \rangle+2d_2d_3^2+v_2=0\\
\frac{\partial f}{\partial d_3} = -2d_1\langle\hat{S}_{13},V_1V_3^{\intercal}    \rangle+2d_1^2d_3-2d_2\langle \hat{S}_{23},V_2V_3^{\intercal} \rangle+2d_2^2d_3+v_3=0
    \end{split}
\end{equation*}

\begin{equation*}
     \begin{split}
d_1\frac{\partial f}{\partial d_1} = -2d_1d_2\langle \hat{S}_{12},V_1V_2^{\intercal}    \rangle+2d_1^2d_2^2-2d_1d_3\langle \hat{S}_{13},V_1V_3^{\intercal} \rangle+2d_1^2d_3^2+d_1v_1=A_1+A_2=0\\
d_2\frac{\partial f}{\partial d_2} = -2d_1d_2\langle\hat{S}_{12},V_1V_2^{\intercal}    \rangle+2d_1^2d_2^2-2d_2d_3\langle \hat{S}_{23},V_2V_3^{\intercal} \rangle+2d_2^2d_3^2+d_2v_2=A_1+A_3=0\\
d_3\frac{\partial f}{\partial d_3} = -2d_1d_3\langle\hat{S}_{13},V_1V_3^{\intercal}    \rangle+2d_1^2d_3^2-2d_2d_3\langle \hat{S}_{23},V_2V_3^{\intercal} \rangle+2d_2^2d_3^2+d_3v_3=A_2+A_3=0
    \end{split}
\end{equation*}

where 
\begin{equation*}
    \begin{split}
        A_1=-2d_1d_2\langle \hat{S}_{12},V_1V_2^{\intercal}    \rangle+2d_1^2d_2^2\\
        A_2=-2d_1d_3\langle \hat{S}_{13},V_1V_3^{\intercal} \rangle+2d_1^2d_3^2\\
        A_3=-2d_2d_3\langle \hat{S}_{23},V_2V_3^{\intercal} \rangle+2d_2^2d_3^2
    \end{split}
\end{equation*}

It is straightforward to derive that $A_1=A_2=A_3=0$, suggesting that 
\begin{equation*}
\begin{split}
    d_1d_2=\langle \hat{S}_{12},V_1V_2^{\intercal}    \rangle\\
    d_1d_3=\langle \hat{S}_{13},V_1V_3^{\intercal} \rangle\\
    d_2d_3=\langle \hat{S}_{23},V_2V_3^{\intercal} \rangle
    \end{split}
\end{equation*}

\begin{equation*}
    f(V_1,V_2,V_3)=-(\langle \hat{S}_{12},V_1V_2^{\intercal}\rangle)^2-(\langle \hat{S}_{13},V_1V_3^{\intercal}\rangle)^2-(\langle \hat{S}_{23},V_2V_3^{\intercal}\rangle)^2
\end{equation*}

\newpage
\bibliography{LinXiao-thesis}

\end{document}